\documentclass{egpubl}
\usepackage{eurovis2021}

\JournalSubmission   

\usepackage[T1]{fontenc}
\usepackage{dfadobe}  

\usepackage{cite}
\BibtexOrBiblatex

\electronicVersion
\PrintedOrElectronic

\ifpdf \usepackage[pdftex]{graphicx} \pdfcompresslevel=9
\else \usepackage[dvips]{graphicx} \fi

\usepackage{egweblnk}

\usepackage{enumitem}
\usepackage{amsmath}
\usepackage{amssymb}

\usepackage{tikz}
\newcommand{\circledw}[1]{\tikz[baseline=(char.base)]{\node[shape=circle, fill=white, draw=black, text=black, inner sep=0.85pt] (char) {\sffamily\footnotesize{#1}};}}

\title[Explainable Adversarial Attacks in DNNs Using Activation Profiles]
      {Explainable Adversarial Attacks in Deep Neural Networks Using Activation Profiles}

\author[G. D. Cantareira, R. F. de Mello \& F. V. Paulovich]
{\parbox{\textwidth}{\centering G.\,D. Cantareira$^{1,3}$,
         R.\,F. Mello$^3$, and
         F.\,V. Paulovich$^{2,3}$
         }
         \\
 {\parbox{\textwidth}{\centering $^1$ King's College London, Department of Informatics, United Kingdom\\
          $^2$ Dalhousie University, Faculty of Computer Science, Canada\\
          $^3$ Universidade de São Paulo, Instituto de Ciências Matemáticas e Computação, Brazil
       } 
 }
 }

\begin{document}

\teaser{
\vspace{-1cm}
 \includegraphics[width=0.95\linewidth]{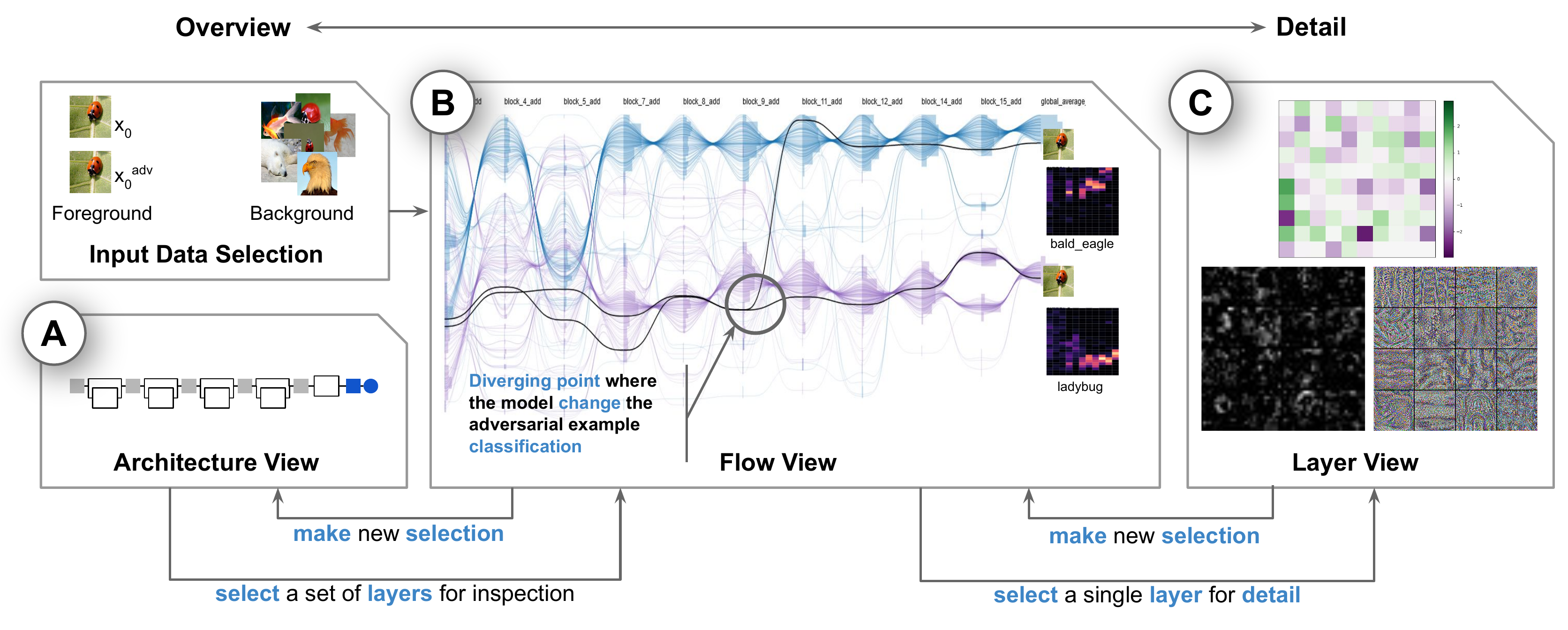}
 \centering
  \caption{Proposed framework overview. Original and adversarial examples are selected as input for a DNN model. Users then use the Architecture View (A) to select sequential layers to visualize. Input examples are then mapped in the Flow View (B) to (black) lines representing the selected layers' responses (outputs). At a certain point, the distance between these lines diverges drastically, indicating the turning point in the model for the adversarial example misclassification. Users can then select the layer in which this change happens for details using the Layer View (C), which shows activation heatmaps and detailed information for each unit in the layer so that the precise units and input features being exploited by an adversarial attack can be pinpointed. It is also possible to return to the previous steps to refine selections and observe different model parts.
 }
\label{fig:teaser}
}

\maketitle

\begin{abstract}
As neural networks become the tool of choice to solve an increasing variety of problems in our society, adversarial attacks become critical. The possibility of generating data instances deliberately designed to fool a network's analysis can have disastrous consequences. Recent work has shown that commonly used methods for model training often result in fragile abstract representations that are particularly vulnerable to such attacks. This paper presents a visual framework to investigate neural network models subjected to adversarial examples, revealing how models' perception of the adversarial data differs from regular data instances and their relationships with class perception. Through different use cases, we show how observing these elements can quickly pinpoint exploited areas in a model, allowing further study of vulnerable features in input data and serving as a guide to improving model training and architecture.

\end{abstract}

\section{Introduction}
\label{sec:introduction}

Artificial Intelligence and Machine Learning are currently among the state-of-the-art technologies to model and analyze complex phenomena, with applications in malware detection~\cite{ye2017malwaresurvey}, robotics and self-driving systems~\cite{grigorescu2020selfdrivingsurvey}, speech recognition~\cite{padmanabhan2015speechsurvey}, biometric authentication~\cite{sundararajan2018biometricsurvey}, among others. A subset of modern machine learning technology, the deep learning models are neural network (NN) that perceive inputs as a deep hierarchy of abstract representations, modeling data beyond most heuristics and finding high-level abstract patterns~\cite{lecun2015deep}.

With the increasing adoption of deep learning models in several systems that surround human life, model security and integrity have become concerning issues~\cite{ma2019vulnerabilities,han2020deep}. An example of a security breach is the \emph{Adversarial Attack}~\cite{chakraborty2018survey}, which consists of an attempt to generate input data (or modify existing data) to fool model interpretation. Adversarial attacks have been employed in almost every scenario that uses machine learning, from writing spam e-mails undetected by spam filters~\cite{nelson2008spam} to forging biometric data that can pass authentication~\cite{biggio2015biometric}. Recent research has shown that deep learning models offer several vulnerabilities to adversarial attacks~\cite{szegedy2013intriguing}. Hence, defending against such attacks has become an important research topic, and many approaches to improve model security and robustness have been proposed, including improvements to model design, training data augmentation, input preprocessing, defensive validation, among others. Identifying vulnerabilities and addressing them also plays a vital role in obtaining a more robust model~\cite{madry2017towards, ma2019vulnerabilities}. 

In this context, Visual Analytics (VA) can aid in developing improved models and better security. Machine learning models are often used as black boxes and benefit from explanation methods, to which VA has offered a sizable amount of solutions~\cite{hohman2018visual}. Visual descriptions of how adversarial examples lead a model to incorrect conclusions may provide essential information on improving security or awareness of existing vulnerabilities. While there are many VA approaches to understanding models and visualizing inner structures, they do not offer a comprehensive view of a model's perception of data, which is essential when comparing adversarial examples against known patterns in the training data distribution. There are approaches explicitly designed for exploring adversarial attacks and vulnerabilities. However, they are either focused on analyzing vulnerabilities in training data distribution~\cite{ma2019vulnerabilities} or lack visual descriptors on how model interpretation evolves in a global context and how hidden layer interpretations of data relate to each other~\cite{liu2018analyzing}.

This paper proposes a VA framework to explain adversarial attacks, focusing on how adversarial data is perceived by a trained model and how model vulnerabilities are exploited on a unit activation level. We explore layers and weights inside the model to determine which areas are triggered by adversarial examples and identify vulnerabilities in input patterns. Our framework allows comparison between adversarial examples and training data, either in single control examples or groups of examples with similar characteristics. This comparison can be made at different levels so that users can pinpoint diverging paths and obtain insights over which areas to explore in detail, identifying patterns to be reinforced in further training or individual units with undesired behavior.

In short, our main contributions are:
\begin{itemize}
    \item A VA framework that explores how adversarial attacks interact with inner elements of a neural network;
    \item Activation Profiles, a similarity-based representation of models perception of data at different hidden layers, allowing for the visual comparison of different data instances; and
    \item Different visual metaphors for condensing groups of activation data, enabling visual understanding of how classes are defined and separated inside a model.
\end{itemize}
\section{Related Work}

\textbf{Visual Analysis of Neural Networks.} Machine learning (ML) models have been widely successful in many different applications over the past years. However, with their continually increasing adoption, a problem has become more pronounced: the black-box nature of such models makes them difficult to interpret, hindering processes such as implementation, improvement, replacement, or adaptation~\cite{hohman2018visual}. Several efforts have been made by the machine learning and visual analytics research communities directed at improving model explainability, comprehending a field known as Explainable Artificial Intelligence (XAI)~\cite{adadi2018peeking}.

Visual explanations for ML models, in particular deep neural networks (DNNs), have different approaches and goals, offering visual representations of various aspects of a model. In a 2018 survey, Hohman et al.~\cite{hohman2018visual} classify existing techniques according to the end goal, visualized data, moment of visualization, intended user, base visual metaphor, and application. Conversely, Liu et al~\cite{liu2017towards} and Choo et al.~\cite{choo2018visual} categorize techniques according to their perceived main goal: model understanding, performance diagnosis, and model refinement. In this context, many techniques fit multiple categories, as the primary goal of XAI approaches is to provide understanding leading to improvements.

Model understanding techniques allow users to investigate a model's analytic process and why it performs in a certain way. They can be model-independent, focusing on broader aspects of neural network functions~\cite{wongsuphasawat2017visualizing, kahng2017activis}, or focused on understanding a specific network architecture, such as Recurrent Neural Networks (RNNs)~\cite{kwon2018retainvis, ming2017understanding, karpathy2015visualizing, li2015visualizing, strobelt2017lstmvis}, Convolutional Neural Networks (CNNs)~\cite{liu2017cnnvis, pezzotti2018deepeyes, liu2018deeptracker, wang2020cnn}, Q-networks (QNNs)~\cite{wang2019dqnviz}, or Generative Adversarial Networks (GANs)~\cite{kahng2018gan,wang2018ganviz}. In common, they present structural views of how the model processes data and how it behaves when subjected to a specific data instance or set of instances. Certain techniques are also directed towards specific data types, as they are often related to the network layout~\cite{nie2018visualizing}. Most visual representations are based on heatmaps, graphs, and multidimensional projections representing weight values, activation outputs, and network layout. Information can be displayed in a weight-centric, data-centric, or neuron-centric fashion, depending on the analysis at hand. Many techniques are mainly defined by how these representations are joined to provide relevant information for a particular task.

Performance diagnosis techniques are designed to help comprehend model training and efficacy at a given task, allowing users to identify problematic data or pinpoint causes of a failed training process that are not easily summarized by metrics such as accuracy or recall~\cite{amershi2015modeltracker}. Model refinement is often achieved by incorporating additional user knowledge to an existing model, and certain techniques are designed to help users identify improvements or fill gaps in areas of uncertainty in input space. Previously mentioned techniques offer debugging tools for interactive refinement and observation of model behavior~\cite{wongsuphasawat2017visualizing, strobelt2017lstmvis}, while others focus on iteratively providing performance explanations through instance or attribute-based representations~\cite{krause2017workflow, krause2014infuse, ren2016squares, zhang2018manifold}. These methods may offer insights when used in the context of adversarial attacks but lack a straightforward way of relating this information to changes observed in final output behavior when analyzing adversarial examples.

A common approach to explaining internal abstract structures in neural networks is to reflect intermediate model outputs into input space. This approach is often associated with image data and CNNs. Several techniques are dedicated to analyzing the distribution of processing unit activations for an input image and generating heatmaps of each pixel's influence in the model's final decision, also known as \emph{saliency maps}~\cite{kindermans2017learning, simonyan2013deep, samek2016evaluating, mahendran2016salient, montavon2017explaining, babiker2017introduction, ancona2017unified, selvaraju2017grad}. Other techniques are focused on optimizing an image to display patterns recognized by the network, such as maximizing layer activations~\cite{yosinski2015understanding, nguyen2016multifaceted}, using another trained model to generate images from intermediate layer outputs\cite{zeiler2014visualizing, zeiler2011deconv}, or applying activations from different images to perform style transfer~\cite{mordvintsev2015incept}. Feature explanation frameworks such as SUMMIT~\cite{hohman2019summit}, delve directly into visualizing layer activations, aiming to correlate classification, descriptive features, and processing unit activations inside hidden layers to understand the meaning of specific activation patterns. In the context of our work, the information provided by these techniques is useful in translating an exploited adversarial vulnerability into patterns in input data. However, as modern models become larger and more complex, identifying which layers or processing units to visualize is not a trivial task. Additionally, showing patterns over input data is not enough when studying adversarial attacks as the origin of such patterns is also a subject to be investigated.

Our framework's primary goal is to generate high-level representations of a model connected to low-level feature explanations to quickly pinpoint features and processing units being exploited by an adversarial example. The concept of visualizing the overall flow of data inside models has been approached by some authors and model understanding techniques~\cite{cantareira2020visualizing, halnaut2020flows, liu2017cnnvis}, but comparisons between data examples with the provided level of information are limited. In particular, Halnaut et al.~\cite{halnaut2020flows} present a method that clusters data instances at each layer output and then displays layers as a sequence of vertical axes, with object exchanges between clusters at each layer shown as a Sankey diagram. However, each layer's visual space is not consistent, and cluster positioning is arbitrary, which would lead to inconsistent representations for classes and make comparisons between instances difficult in our scenario.

\textbf{Adversarial Attacks and Visual Analytics.} An \emph{adversarial attack} is a manipulation of input data to force a trained ML model to produce incorrect outputs~\cite{chakraborty2018survey}. In the context of neural networks, Szegedy et al.~\cite{szegedy2013intriguing} discovered that several state-of-the-art NN models were vulnerable to \emph{adversarial examples}, i.e., manipulated input data that would be misclassified by the model, despite being largely similar to correctly classified examples drawn from the data distribution. With the adoption of deep learning in an increasing variety of applications, the subject of adversarial attacks has drawn the research community's attention, with many recent developments in both building models with better security and producing more sophisticated attacks.

Adversarial machine learning literature describes four main features of an adversary: goal, knowledge, capability, and strategy~\cite{biggio2018wild}. An attack is often described by its operation regarding one of these features. \textit{Black-box} vs. \textit{white-box} attacks, for instance, refer to adversary knowledge of model details. Attacks can be conducted before or after a model is trained, depending on the adversary's goal and access capability: \textit{Poisoning} attacks manipulate training data to induce a particular behavior in the optimized model, such as misclassifying examples~\cite{munoz2017towards} or providing a backdoor when a specific pattern is encountered~\cite{gao2019strip}. Conversely, \textit{evasion} attacks target an already trained model during run time, aiming to force an incorrect output~\cite{kwon2018multi}.

Recent research shows sophisticated approaches to producing adversarial examples, from methods to add undetectable triggers in training images~\cite{saha2019hidden, liao2018backdoor} to generating misclassified examples by altering a single pixel~\cite{su2019one}. Efforts in explaining the causes behind a model's vulnerability have also been made, such as analyzing robustness concerning output functions~\cite{carlini2017towards, ozbulak2018softmax, rozsa2017robustness}, or showing how adversarial vulnerabilities can reveal insights on training data~\cite{ilyas2019adversarial}. 

Currently, literature comprehending visual methods to interpret and understand the causes and effects of adversarial attacks in neural network models is quite limited. Ma et al.~\cite{ma2019vulnerabilities} proposed a framework to explain classifier vulnerabilities through high-dimensional relationships between training data examples. However, this framework does not provide information regarding how such vulnerabilities relate to abstract representations formed inside a model, which may be vital in determining model trustworthiness~\cite{ribeiro2016trust}. Liu et al.~\cite{liu2018analyzing,cao2020analyzing} proposed AEVis, a VA tool that explores models perception of adversarial examples by extracting paths taken by data inside a network and allowing comparison in multiple levels (layer, feature map, neuron). While the multi-level aspect of this approach is informative and in ways similar to what we present in our framework, it has some limitations. Although datapaths can identify the most critical units towards a particular classification at each layer, guidance as to which layers may be of interest (namely, \textit{diverging points} where the model perception of data instances changes drastically) is limited, as it only shows cosine dissimilarity between outputs at each layer using dots. Additionally, its explanation process is focused on the observation of robust features, and understanding non-robust features may be crucial when investigating adversarial examples~\cite{ilyas2019adversarial}.

\section{Design Goals}

With the problem at hand and limitations of existing approaches in mind, we devised a framework to explain adversarial attacks in an internal representation context through VA. While the concepts presented here can be employed in other applications, our framework is based on CNN models for image classification, which is a common subject in literature~\cite{krizhevsky12, krizhevsky2010}. We also focus our discussion on \textit{evasion attacks}, i.e., when the attacker's goal is assumed to be manipulating an image to provoke incorrect classification. In this paper, adversarial examples are generated with full knowledge of the model (white-box perspective).

After reviewing the literature on evasion attacks~\cite{yuan2019adversarial,xu2019security,akhtar2018threat,jiang2020attacks,kwon2018multi} and adversarial image generation~\cite{goodfellow2014explaining,madry2017towards,athalye2018synthesizing}, how they relate with XAI approaches in VA, as well as discussing with an ML expert, we defined the following tasks to be performed by our framework:

\begin{itemize}
    \item[T1.] \textbf{Summarize Model Perception of Adversarial Examples.} Our goal is to provide an understanding of how an adversarial example exploits the inner properties of a neural network. To achieve this, our first step is to provide a global view of hidden outputs so that the user can quickly grasp model behavior when exposed to a given input, as well as perform comparisons between adversarial and regular data examples so that points of abnormal behavior can be quickly identified and further examined.
    \item[T2.] \textbf{Identify Vulnerable Features.} Once an abnormal behavior point is chosen, we must be able to delve into the model's processing units to identify which features or filters were the most responsible for the observed difference, and therefore the most exploited by the attack. The user can then visualize these features to correlate them with their expertise or identify robust vs. non-robust features.
\end{itemize}

From the analytical tasks, we devise a set of VA requirements to be fulfilled:

\begin{itemize}
    \item[R1.] \textbf{Highlight areas of interest.} There are visualization techniques focused on providing an understanding of layers or features, but visiting each layer and feature is impractical (visually and computationally) when dealing with models with a large number of layers or processing units per layer. Therefore, a way to profile global data flow for a given input to identify and explore where pivotal changes happen is needed. 
    \item[R2.] \textbf{Visual comparison between data flows.} Besides being global, data flow through hidden outputs must also be comparable in a simple manner. It is necessary to understand how the model perceives regular data and adversarial examples and pinpoint locations where divergences are present. 
    \item[R3.] \textbf{Visual profile for data groups.} When observing data flow in a classification model, it also becomes necessary to represent output patterns from multiple input data, such as classes. To achieve this, we need a visualization for the accumulation of outputs from several data instances, displaying taken paths and density.
    \item[R4.] \textbf{Feature exploration.} Once we can pinpoint a layer where substantial differences occur, we then need to understand how they came to be. A detailed examination of the actual meaning of features and their impact on data separation is needed. 
\end{itemize}

\section{Framework Structure}

\subsection{Overview}

Our framework is divided into three main components, presented in Figure~\ref{fig:teaser}. The first is the \textit{Architecture View} \circledw{A}, which displays the model under analysis topology as a compact graph representation acting as the starting point where users select a sequence of layers to display. The second is the \textit{Flow View} \circledw{B}, which shows data objects progressing through the model's selected layers. These data objects can be displayed individually or in groups. The selection of layers displayed at the \textit{Flow View} can be changed at any moment, with the possibility of selecting a set of representative layers from the full extension of the model, visualizing them to find an area of interest, and then making a new selection with layers from that area to explore in more detail (tasks \textbf{T1}, \textbf{T2}). The last component is the \textit{Layer View} \circledw{C}, in which users can choose a single layer to explore on a per-unit basis. Here, users can observe outputs from every unit in the layer, identifying patterns for groups, or comparing outputs between individual data objects (task \textbf{T2}).

\vspace{-0.35cm}
\subsection{Architecture View}
\label{sec:topology}

To help users select which layers to visualize, we first present a compact graph representation of the entire model, similar to the computation graph proposed in~\cite{kahng2017activis} or TensorBoard's graph view (\url{https://www.tensorflow.org/tensorboard}). This graph is organized horizontally from input to output layers. Layers with external inputs or outputs are represented as circles, while other layers are represented as squares linked to their inputs on the left side and outputs on the right side. They are color-coded according to type: layers that do not have any trainable parameters are colored in gray. Connected layers with trainable weights (e.g., dense or convolutional layers) are colored in blue, and other trainable layers (e.g., batch normalization) are colored in red. Multiple layers can be grouped in blocks, which are shown as white rectangles. Branching connections (layers with multiple input or output tensors) between blocks are kept visible since a selection of layers that is non-sequential should not be displayed at the \textit{Flow View}. This representation is designed to occupy as little space as possible while providing models overview.
 
\vspace{-0.35cm}
\subsection{Flow View}

An essential task in our framework is to provide a representation of models perception of adversarial data to find points of interest to zoom in further and explore in detail (task \textbf{T1}). Highlighting layers that significantly improve an example's similarity towards a target group or further separate it from its original class accelerates visual analysis and avoids users having to explore a model on a layer-by-layer basis (requirement \textbf{R1}). This task can be performed using the \textit{Flow View} component, which offers visual representations to support layers' interpretation given an input data set.

\begin{figure}[!ht]
    \centering
    \includegraphics[width=\columnwidth]{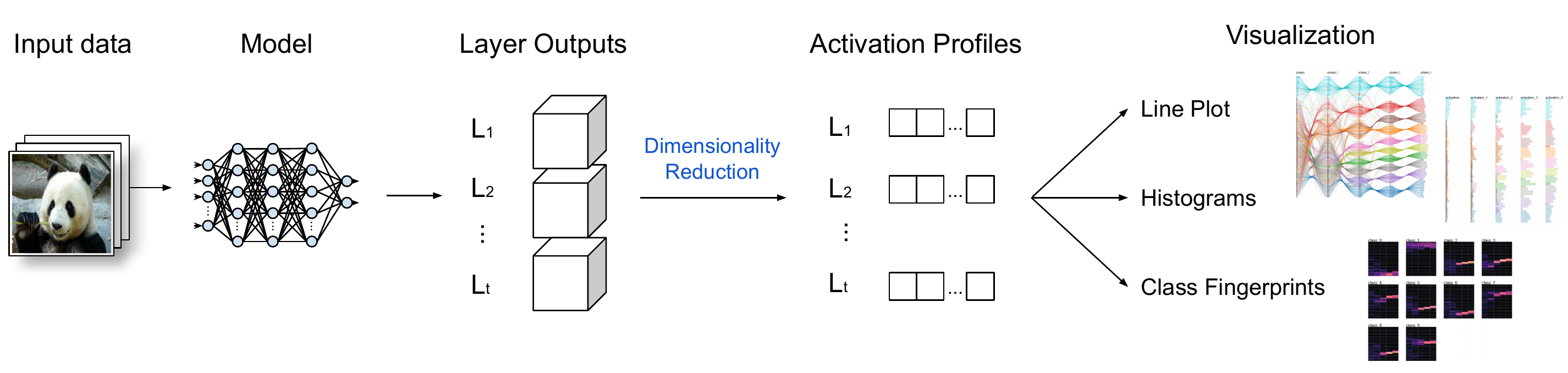}
    \caption{\textit{Flow View} overview. Input data is processed by the model and hidden outputs of selected layers are transformed into 1-D arrays using a dimensionality reduction approach, resulting in \textit{Activation Profiles}. \textit{Activation Profiles} are then used to build visualizations, enabling the comparisons of regular and adversarial examples considering the model's perception.
    }
    \label{fig:context_diag_0}
\end{figure}

Once a set of layers $L = (L_1, L_2,..., L_t)$ is selected at the  \textit{Architecture View}, input data is processed through the model, and outputs at each layer are displayed. In our framework, we analyze a set of input data $X$ considering two different perspectives: \textit{background} and \textit{foreground} data. Background data provides context to the visualization, being displayed as groups (requirement \textbf{R3}). Foreground data are objects to be displayed individually: adversarial and control examples (requirement \textbf{R2}).

To show how different layers perceive data objects and how this perception evolves as data travels further in the model, we build \textit{Activation Profiles}. For every layer $L_i \in L$ and data object $x_j \in X$, hidden outputs $L_i(x_j)$ are transformed into scalar values $p_i(x_j)$ whose distances in $\mathbb{R}^1$ space represent dissimilarity relationships in $L_i(X)$ as best as possible. For each $x_j$, then, there is an 1-D array $p(x_j) = [p_1(x_j),p_2(x_j),...,p_n(x_j)]$ containing its interpretation according to each layer in $L$. As stated in Sec.~\ref{sec:topology}, layer selection should follow a sequential pattern, so that outputs from $L_i$ are in some way part of the input in $L_{i+1}$ even if they are not immediately connected. Non-sequential models can be explored using the framework, as long as the selected layers are sequential.

To transform multivariate layer outputs into scalar values, we employ UMAP~\cite{mcinnes2018umap}. Although the information contained in hidden layer outputs cannot be fully represented in one dimension, full distance preservation is not our goal. We guide the projection process to focus on displaying distance relations that change between layer outputs, preserving relative cluster positions, and providing relevant information for comparison. This is achieved by applying an alignment constraint~\cite{cantareira2020generic} to all projections. This constraint ensures that overall cluster positioning is similar in all projected layers, providing context for changes between projections (requirement \textbf{R1}). Additionally, alignment indirectly favors accumulating error in the same pairwise distances across all projections, meaning that if a distance between two points cannot be properly replicated but does not display expressive change across projections, its representation will remain relatively stable.

After \textit{Activation Profiles} $p(X)$ values are obtained, we build the \textit{Flow View}. Foreground data are visualized directly: selected layers are shown as a sequence of vertical axes, and each data object is represented as a line connecting its positions at each axis, similar to a Parallel Coordinates~\cite{heinrich2013pc} visualization. Background data are shown in the same way, but lines are colored according to label (either true or predicted) and curved and bundled together~\cite{zhou2013edge} using label and initial and final positions between each pair of layers as clustering inputs in the bundling process. This representation, which we call \textit{Background Flow}, allows for direct comparison between flow patterns from different groups (requirement \textbf{R3}) and offers a non-intrusive background to provide context when comparing two individual instances (requirement \textbf{R2}).

Background context is essential to providing meaning to object positioning. When comparing only two data objects, the default approach measures and shows distances between them at each layer output~\cite{liu2018analyzing}. However, layer outputs may vary, both in the number of dimensions and magnitude (especially when considering factors such as vanishing or exploding gradients), and correctly normalizing values to enable proper comparison may become a problem. Additionally, only stating that objects became closer or further apart is not enough -- by examining them in contrast to background groups, it is possible to notice if such distancing is atypical, or if an object became closer to a different group (requirements \textbf{R2} and  \textbf{R3}). In our framework, background and foreground data are projected as a single dataset. UMAP is designed so that a projection can be fit to background data and then used to transform future data observations, allowing exploration and comparison of multiple examples while maintaining the same context (requirement \textbf{R3}).

In addition to \textit{background flow}, we devise two more visualizations for background data to complement the representation of classes and other groupings. The first, \textit{profile histogram}, is composed of histograms over each layer axis, providing a clear notion of density at different parts of a line. The second consists of joining all histograms for each class in a heatmap providing a compact visual descriptor of each class that can be compared side-by-side, which we call \textit{Class Fingerprint} (requirements \textbf{R2} and \textbf{R3}).

Figure~\ref{fig:context_diag_0} shows the process for visualizing data in the \textit{Flow View}. Using this view, users can iteratively select different layer combinations to observe how a model's perception of objects evolve and pinpoint changes in behavior. Once a particular layer of interest is found, it can be displayed using the \textit{Layer View}.

\vspace{-0.35cm}
\subsection{Layer View}

\begin{figure}[!ht]
    \centering
    \includegraphics[width=\columnwidth]{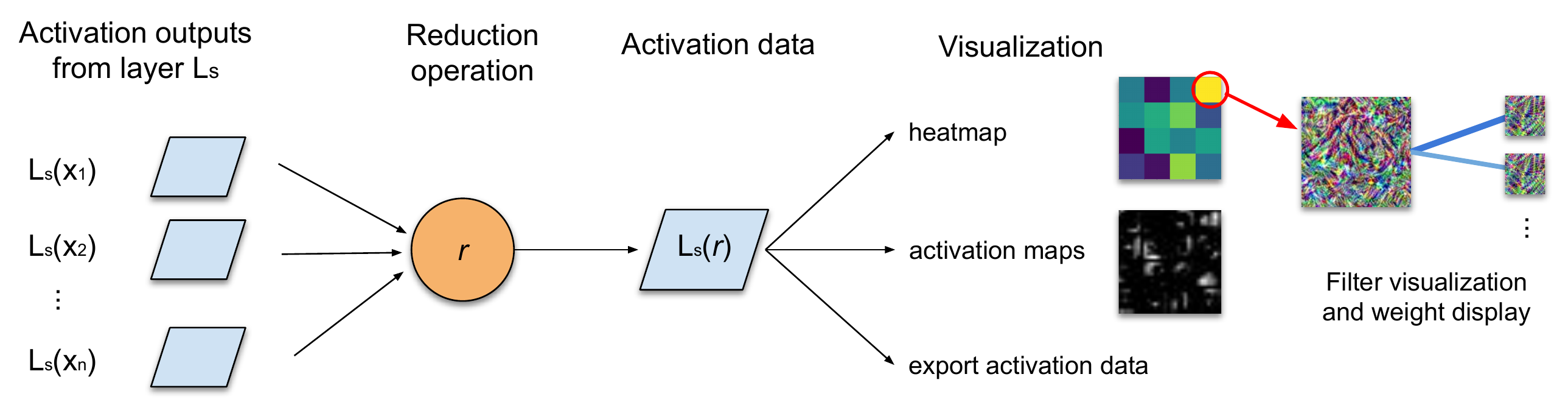}
    \caption{\textit{Layer View} overview. Outputs from a selected layer are subjected to reduction to obtain a single output equivalent, representing activation data. This data is displayed using heatmap visualizations to highlight units with interesting behavior. Units can be selected and displayed using convolutional filter visualization and highlighting weights in their connections. Activation data can also be exported for further statistical analysis.}
    \label{fig:detail_diag_1}
\end{figure}

The \textit{Layer View} is focused on displaying output information $L_s(x)$ from a single layer $L_s$, selected in the \textit{Flow View}, to allow users to investigate the causes behind changes in behavior, down to individual unit (filter) activations. Visual metaphors employed in the \textit{Layer View} merge and extend ideas presented in~\cite{kahng2017activis, hohman2019summit}. Analog to the \textit{Flow View}, outputs are obtained from a set of input examples $X$; either foreground or background data can be selected for visualization. Units can be selected and inspected individually, through weight display and filter visualization~\cite{yosinski2015understanding}. Figure~\ref{fig:detail_diag_1} illustrates this process.

\vspace{-0.2cm}

The \textit{Layer View} produces visual representations from outputs $L_s(x_i)$ from a single object $x_i$ at a time. We formulate it so that the output set $L_s(X)$ is subjected to one of four reduction operations, producing the activation data $L_s(r)$, which has the same dimensions as an output for a single data instance. The choice of operation depends on the exploratory task to be conducted:

\begin{itemize}
    \item \textbf{mean$()$}: corresponds to the mean output values at each unit for the outputs $L_s(X)$, which we use to determine activation patterns for classes and large groups.
    \item \textbf{max$()$}: maximum values in $L_s(X)$ at each processing unit in $L_s$. This output is used for examining layer-class combinations where small subsets of $X$ produce high outputs in different units spread throughout the layer.
    \item \textbf{difference$(i,j)$}: output with unit-wise difference $(L_s(x_i) - L_s(x_j))$ between two outputs. This is the operation we use to compare outputs between an adversarial image and its original counterpart, as the resulting output contains the units most affected by the adversarial manipulation.
    \item \textbf{direct\_output$(i)$}: allows the observation of a single given output $L_s(x_i)$.
\end{itemize}

The \textit{Layer View} is then composed as a heatmap. All units are arranged in a square grid and colored according to their output value (in the case of 2D convolutional units, we pick the maximum activation from each matrix). Besides the heatmap, activation maps for convolutional units are also shown, consisting of a grid of output images for units in $L_s$ that show the highest activation values, indicating areas that generated the strongest activations in the layer.

Filters can then be selected and visualized through activation maximization~\cite{erhan2009visualizing}. An input image with visual patterns that stimulate filter $L_{s,k}$ is generated from random noise and displayed. Similar images are also generated for filters in $L_{s-1}$, $L_{s+1}$ whose connections to $L_{s,k}$ have the strongest weights, represented as lines connecting them. In this way, users can grasp which abstract representations are the most influential in activating $L_{s,k}$, and which representations are the most influenced by $L_{s,k}$, respectively. Additionally, we can export the generated activation data to perform further analysis, such as building graphs or obtaining more statistical information (ranking units by output, identifying maximum, minimum, and standard deviation in outputs, inspecting covariance between units for multiple outputs from background data, among others).

\section{Study Cases}

In this section, we demonstrate the application of our framework in two study cases. The first case presents an evasion adversarial attack study over a small CNN architecture for classifying MNIST handwritten digits~\cite{lecun1998mnist}. The second shows vulnerable features and units in a deep model, the MobileNet V2~\cite{sandler2018mobilenetv2} trained on ImageNet~\cite{deng09imagenet} data. The framework front-end was implemented using d3.js (\url{https://d3js.org/}), with tensorflow (\url{https://tensorflow.org/}) and keras~(\url{https://keras.io/}) as back-end.

\vspace{-0.35cm}
\subsection{Adversarial Data Generation}

Adversarial machine learning literature defines an adversarial example $x^{adv}_i$ as the result of applying a perturbation $s$ (based on data distribution) over an example $x_i$  to obtain an incorrect outcome from a model.  For image classification purposes, perturbations come from a set of allowed perturbations $S$ that is commonly chosen to ensure that modified examples are still perceptually similar to the originals, such as being bound to an $l_2$ or $l_\infty$ ball around original examples.

In this paper, we use Projected Gradient Descent (PGD)~\cite{madry2017towards} to generate adversarial data. A gradient-based method originating from \textit{Fast Gradient Sign} (FGSM)~\cite{goodfellow2014explaining}, PGD is considered a universal first-order adversary, i.e., its output tends towards the strongest possible attack with local first-order information from a network model. Several studies propose adversarial PGD in training to improve model robustness~\cite{madry2017towards, araujo2019robust, shafahi2019free, wong2020fast}, with remarkable results. We note that other methods for adversarial data generation~\cite{carlini2017adversarial, xiao2018generating, moosavi2016deepfool} could provide their own set of properties to be visualized and explained as well.

\vspace{-0.35cm}
\subsection{Explaining Adversarial Examples}
\label{sec:cnn_mnist_1}

Our first study case consists of the interpretation and exploration of evasion attacks on a small convolutional model. The model is a $4$-layer CNN, trained for image classification on the MNIST handwritten digit dataset. Each layer has a $3x3$ filter, ReLU activation, followed by a max-pooling layer of stride size $2$, effectively halving output dimensions. The number of units at each layer is $4$, $8$, $16$, and $32$. Following the last convolutional layer, a global max-pooling layer is linked directly to a $10$-unit softmax output. 

The training was conducted using Adam optimization (initial learning rate = $0.001$), and accuracy results reached $99$\% for training and test sets after $10$ epochs.

\begin{figure}[!ht]
    \centering
    \includegraphics[width=0.475\textwidth]{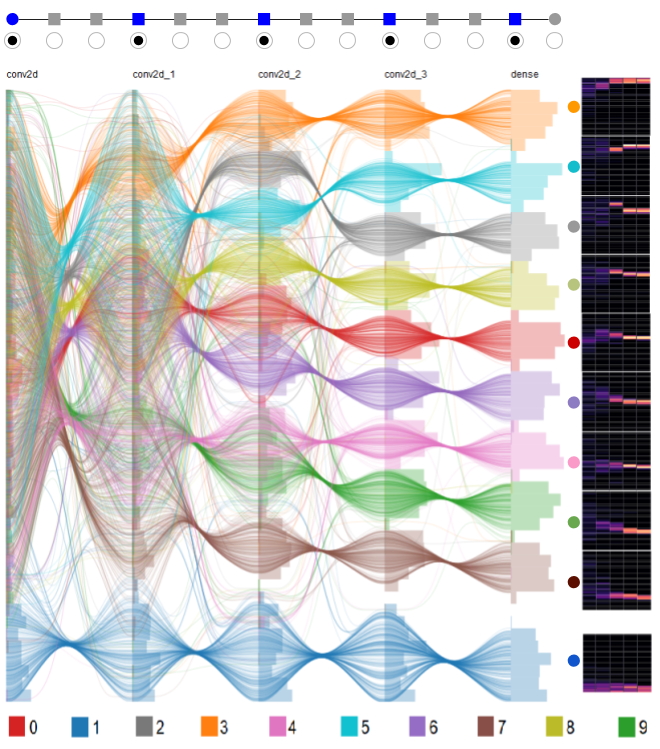}
    \caption{Background data visualization for the CNN-MNIST case study. The upper diagram shows the Architecture View for the model, indicating which layers are selected for view. The line plot shows lines for all background data objects, colored by their predicted label. Class separation may occur at different layers. Many objects in classes ``9'' and ``4'' (brown and orange, respectively) only get properly separated at the fourth convolutional layer, while the second layer already separates many objects in class ``1'' (cyan). Class fingerprints, shown on the right, display the accumulation of points at each segment of the visual space, highlighting condensed areas with brighter colors.}

    \label{fig:mnist_cnn_baseline}
\end{figure}

A $2,000$-object sample $X$ from the training data set was drawn at random as background data to establish a baseline for projections. \textit{Activation Profiles} from this sample in each of the weighted layers (four convolutional, one dense for output - captured before softmax activation) are shown in Figure~\ref{fig:mnist_cnn_baseline}. The \textit{Architecture View} is shown on the upper area, allowing selection of layers to be observed, while the \textit{Flow View} provides information on how each layer affects relationships and separation between different classes. Each class's paths are condensed using \textit{Class Fingerprints}, shown on the right side. In this Figure, it is possible to notice that separation between class groups may occur at different points of the model, depending on the class. For instance, classes  ``9'' and ``4'' only get properly separated at layer $conv2d\_3$, while class ``1'' has many of its instances already grouped at layer $conv2d\_1$. Class fingerprints get brighter as instances are condensed in a region of the vertical axis, indicating higher separation. An adversarial set $X^{adv}$ was created by producing an adversarial image from each object in $X$. Adversarial data were generated using PGD, with parameters $\epsilon = 0.3$, $\alpha = 0.075$ and $100$ iterations. When predicted by the model, accuracy for the adversarial sample was $1.6\%$.

\begin{figure*}[t]
    \centering
    \includegraphics[width=0.9\textwidth]{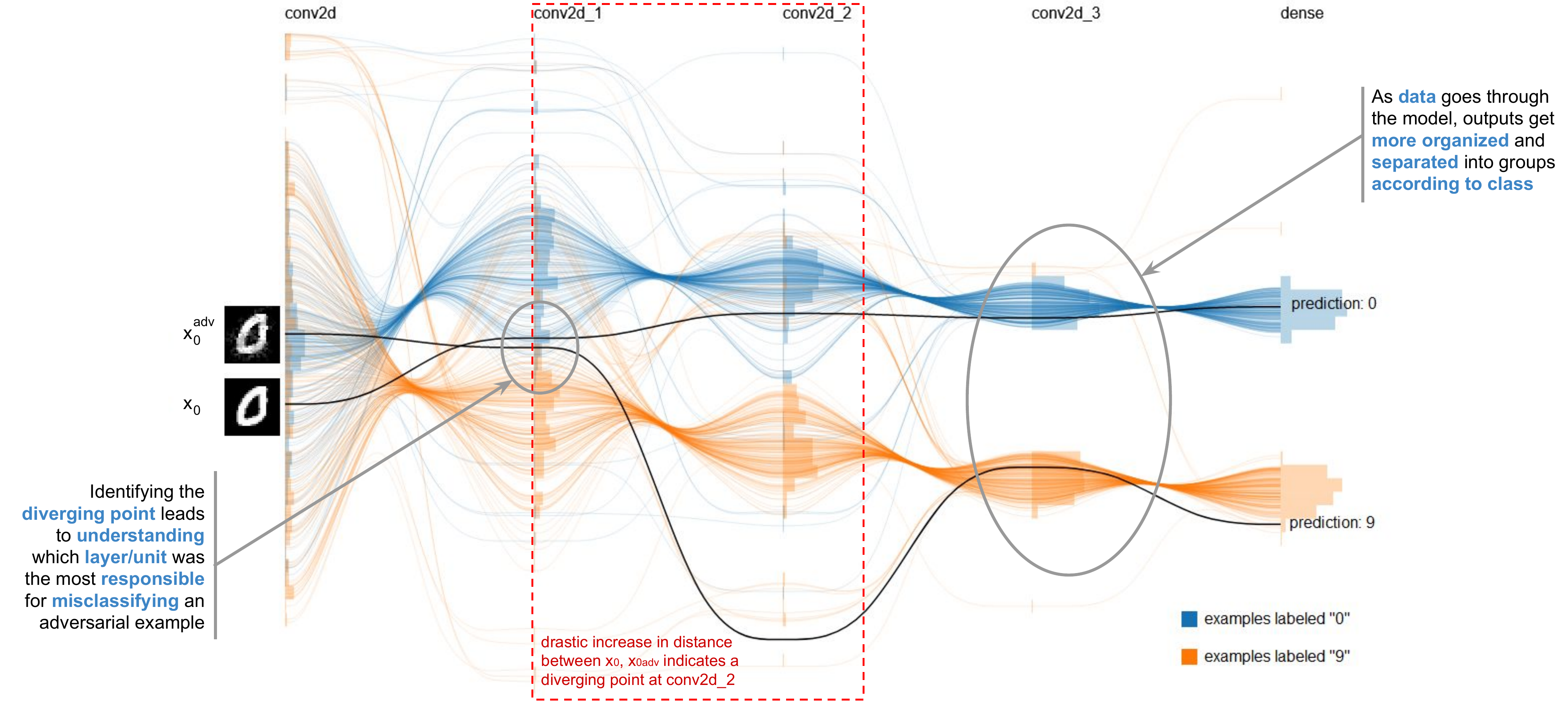}
    \caption{Lines representing original image $x_0$ and adversarial image $x^{adv}_0$ being interpreted by the CNN-MNIST model. Background data representing classes ``0'' and ``9'' is also shown. The two images are perceived as relatively similar until layer $conv2d_2$, when separation occurs and each example merges into a different class. This \textit{diverging point} highlights layer $conv2d_2$ as a point of interest.}
    \label{fig:mnist_cnn_0}
\end{figure*}

Next, we pick an image $x_0$ from $X$ to explore the model's reaction. We run it through the model and project its outputs on the flow view along with its adversarial counterpart, $x^{adv}_0$, and observe how their behaviors change at each layer. Figure~\ref{fig:mnist_cnn_0} shows this comparison. $x_0$ had its label (``0'') correctly predicted by the model, but $x^{adv}_0$ was predicted as belonging to the class ``9''. Both examples are projected alongside background data from their perceived and correct classes. It is possible to observe that, while similar up to the second convolutional layer ($conv2d\_1$), $x_0$ and  $x^{adv}_0$ becomes vastly different afterward. 

At the fourth convolutional layer ($conv2d\_3$), $x^{adv}_0$ is already grouped with other objects in class ``9''. As highlighted in Figure~\ref{fig:mnist_cnn_0}, There is a \textit{diverging point} between $conv2d\_1$ and $conv2d\_2$ where the relationship between both outputs changes drastically, and it is likely to be where units most affected by the adversarial manipulation in $x^{adv}_0$ are located. This behavior was observed in all experiments we performed; while output similarity between a given $x_i$ and its adversarial deviation $x^{adv}_i$ may vary at the first layers of a model due to perturbations being detected by low-level units, activations at the middle sections see the objects as somewhat similar. When the diverging point is reached, there is a sudden shift, where the two objects become completely separated, and the adversarial example assumes behavior more similar to the perceived (incorrect) class. Having observed where this shift is located, we can proceed to examine layers at both ends of the diverging point in detail.

\begin{figure}[!ht]
    \centering
    \includegraphics[width=0.5\textwidth]{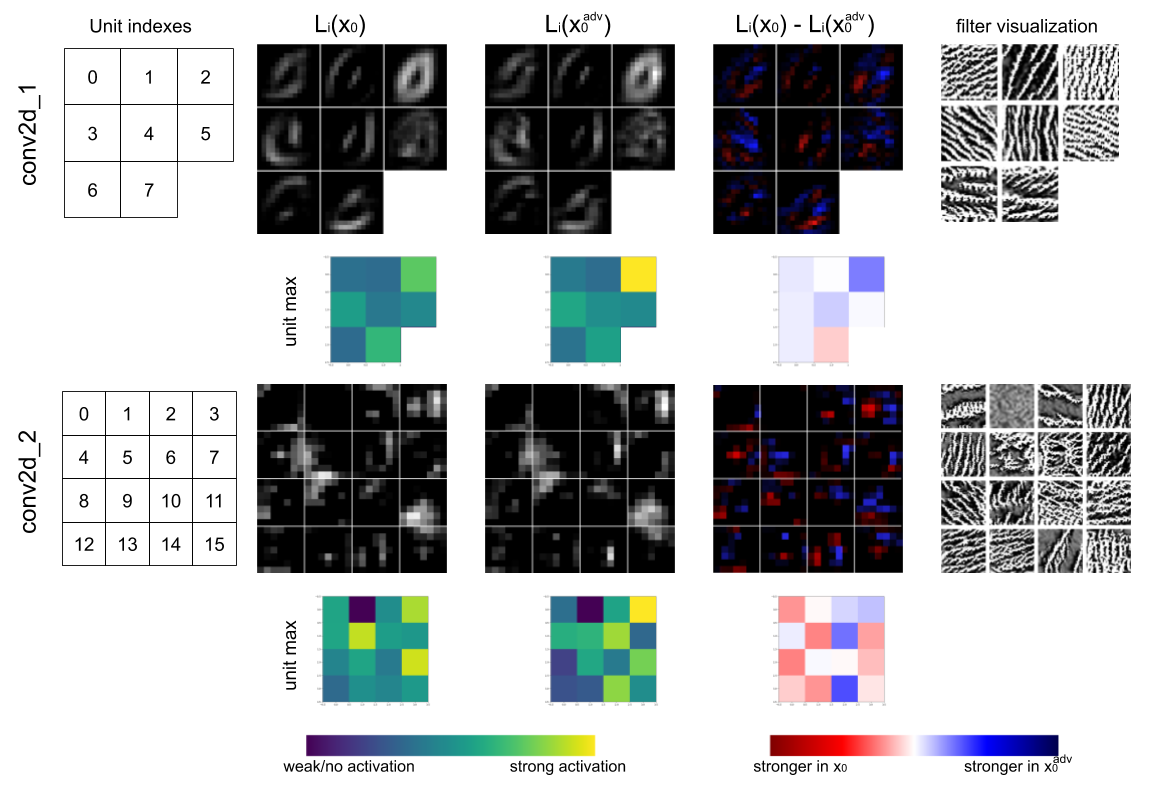}
    \caption{Visualization for layers $conv2d\_1$ and $conv2d\_2$ of the CNN-MNIST model. The first column identifies each unit in the layer. Second and third columns show activations for $x_0$ and $x^{adv}_0$ from the two observed layers, and the fourth column shows the difference in activations between the two. Finally, the fifth column visualizes all filters in each layer using activation maximization. Activations from $x_0$ and $x^{adv}_0$ at the first layer are still relatively similar but show much more variation in the second. Many activations appear to be stronger on the center-right side for $x^{adv}_0$, and stronger on the lower-left side for $x_0$.}
    \label{fig:mnist_cnn_1}
\end{figure}

As this model is narrow and shallow, it is possible to observe all unit outputs for comparison. Figure~\ref{fig:mnist_cnn_1} shows outputs at layers $conv2d\_1$ and $conv2d\_2$ from both images. For each layer, there are two rows. The top row contains complete convolutional layer outputs, with one resulting image for each unit, showing how strong units reacted at each pixel. The bottom row shows \textit{unit maximum} heatmaps colored with the strongest activation among all pixels in that unit. The third column is the difference between activations from $x_0$ and $x^{adv}_0$. Activations that are stronger in $x_0$ are colored in red, while activations stronger in $x^{adv}_0$ are colored in blue. The fourth column shows a visualization of input patterns that activate each unit in that layer.

The layer that precedes the large shift in $x^{adv}_0$'s positioning, $conv2d\_1$ (top row), shows largely similar maximum outputs among all units, with only one of them (top right) showing expressive differences. The unit maximum heatmaps from $conv2d\_2$, however, show many differences between outputs from the two examples. Looking at the activation maps for $L_i(x_0) - L_i(x^{adv}_0)$, it is possible to notice that, in $conv2d\_2$, units that show stronger activations for $x_0$ seem to be concentrated at the lower left side of the image, while stronger activations for $x^{adv}_0$ are generally located on the center-right side. Analyzing the filter visualizations (fourth column), $x^{adv}_0$ appears to be increasing the response of vertical ramification patterns on the right side of the image, which favors label ``9'' while decreasing response on curved patterns on the lower left side of the image ($conv2d\_2$, units $0$ and $13$), which reduces the odds of a label ``0''.

\begin{figure}[!ht]
    \centering
    \includegraphics[width=0.35\textwidth]{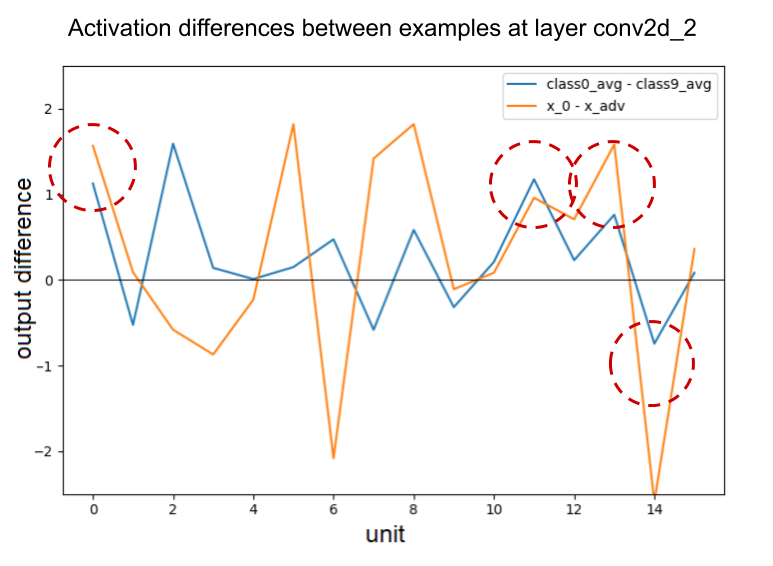}
    \caption{This graph shows the difference between outputs at each unit, for the two examples as well as the two relevant classes. The adversarial example accentuates certain differences in activation (highlighted in red), which can be translated into visual features in Figure~\ref{fig:mnist_cnn_1}, but there are many large differences in activation that do not match class patterns.}
    \label{fig:mnist_cnn_3}
\end{figure}

In additional exploration, in Figure~\ref{fig:mnist_cnn_3}, we export activation data and plot the difference between outputs from the two examples at each unit and the differences between mean activations for background data in classes ``0'' and ``9''. Our goal is to identify units in which differences in activation between classes ``0'' and ``9'' were most reflected in our adversarial example. We can identify four units ($0$, $11$, $13$, and $14$) where changes in $x_{adv}$ follow the same pattern as in class activations. These are not the units with the highest mean activations for each class, representing secondary patterns. Interestingly, the adversarial example also generates highly different activations in certain units (such as $5$,$6$, and $7$) that seem to have no impact in changing predictions between the two classes. 

\vspace{-0.35cm}
\subsection{Vulnerable Features on Deep Networks}
\label{sec:mnet_imagenet_1}

In this case study, we explore hidden layer outputs of a deeper model, locate its diverging point for an adversarial example, and observe how visual features relate to its layer's units. The studied model is \textit{MobileNet V2}, an architecture designed for image data analysis on mobile devices. Keras \textit{MobileNet V2} implementation has $3,538,984$ parameters and $157$ layers. Our model was trained on \textit{ImageNet} images for data classification into $1,000$ different classes, with the output layer positioned immediately after a global average pool of the last convolutional layer.

\begin{figure*}[!h]
\centering
    \includegraphics[width=.8\textwidth]{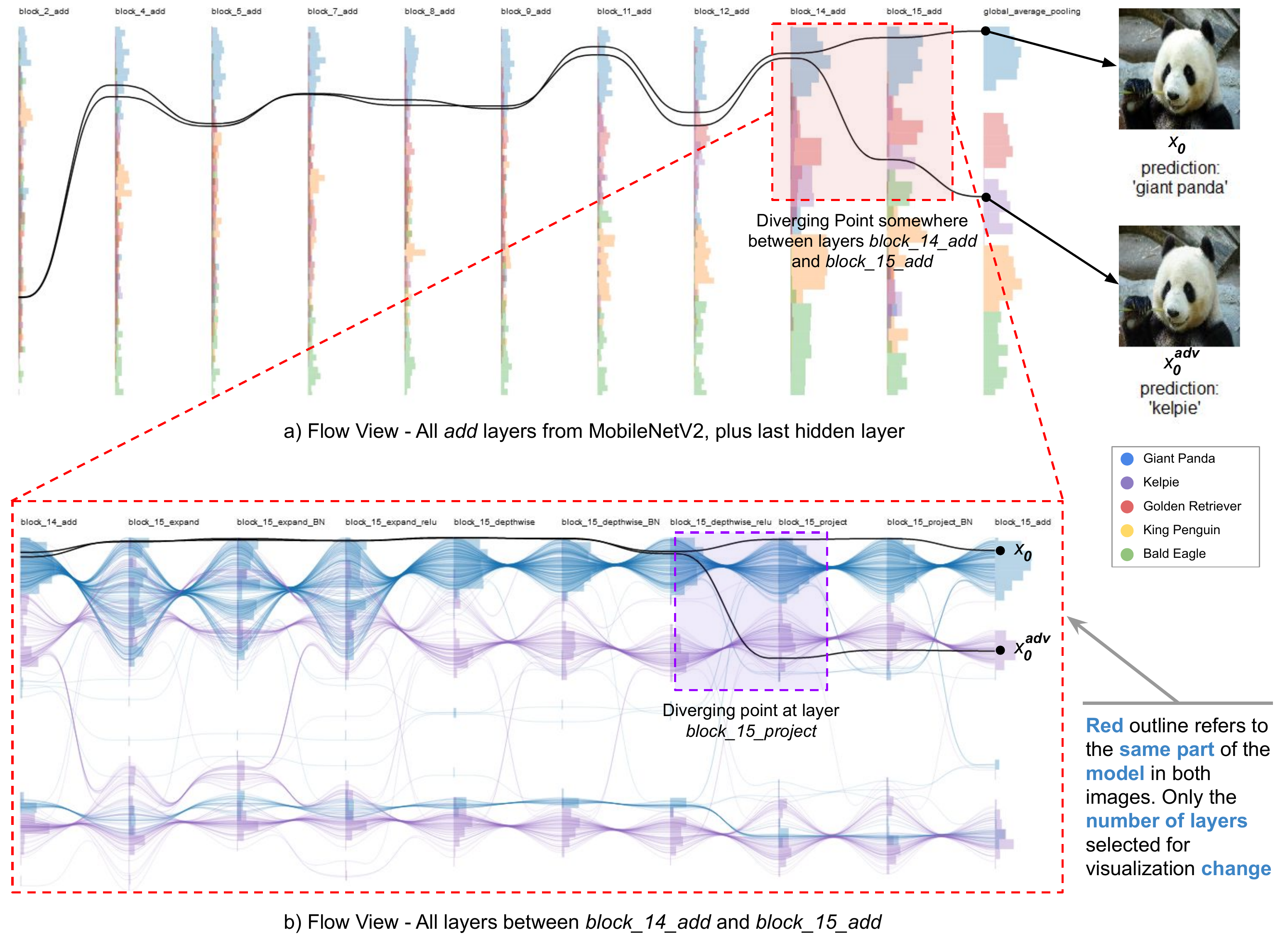}
    \caption{\textit{Flow View} displaying data flow from two images (control example $x_0$ and adversarial example $x^{adv}_0$) plus background data. The upper image contains layers from several parts of the model. A diverging point can be observed in the area highlighted in red, as the two trajectories suddenly branch in different directions between layers $block\_14\_add$ and $block\_15\_add$. All layers contained between these two can then be selected for display, resulting in the lower image. The flow visualization allows investigation and discovery of the exact moment when the two images' model perception splits apart.
}
    \label{fig:imagenet_mobilenet_1}
\end{figure*}

Base image $x_0$ for this study consists of a giant panda photo from \textit{ImageNet}, as seen in Figure~\ref{fig:imagenet_mobilenet_1}. The model is capable of predicting its label correctly, with $93$\% confidence. From this image, we generate an adversarial example $x^{adv}_0$ using PGD, with $\epsilon = 0.3$, $\alpha = 0.075$ and $200$ iterations. When analyzed by the model, $x^{adv}_0$ was misclassified as 'kelpie,' with $99$\% confidence. We use our framework to compare how both objects are perceived and identify where pivotal differences begin to occur.

Since the model is too deep for all layers to be displayed at once in the \textit{Flow View}, we use the \textit{Architecture View} to make a selection of layers of interest. \textit{MobileNet V2} is divided into blocks and has residual components that merge at each block's end. We choose to visualize all $add$ layers in the model ($10$ in total), as they are sequential and evenly spaced enough to provide an approximation for the whole architecture. We also visualize the last hidden output in the network (global average pooling before the dense output).

Background data is also essential for the \textit{Flow View} visualization. Since it will provide context for the two examples being compared, the choice of background data must contain information relevant to the analysis at hand. While it is possible to draw a sample from all $1,000$ classes in \textit{ImageNet}, a large amount of space would be occupied by objects and features that are not relevant for this comparison. On the other hand, using only data from the two involved classes would not provide enough context of what separation between objects might mean. Selecting data with known shared characteristics may lead to important details being highlighted. Therefore, we opt to use $2,000$ examples from five different animal-related classes as background data: 'king penguin', 'eagle', 'golden retriever', 'giant panda', and 'kelpie'.

Figure~\ref{fig:imagenet_mobilenet_1} shows the projected output values for both examples and background data at each of the $11$ selected layers. Here, we opt to show projected background data only as \textit{profile histograms} colored by predicted class for a cleaner representation. Outputs from the two images are considerably similar in the first half of the model. This image's diverging point is easily identifiable: at a certain point between layers $block\_14\_add$ and $block\_15\_add$, model perception of $x_0$ and $x^{adv}_0$ becomes drastically different (area highlighted in red).

In \textit{MobileNet V2}, block $15$ is a sequence of layers that branch off the output from block $14$. At the end of the block, the layer $block\_15\_add$ merges its output back with the unmodified output from block $14$. Since $x_0$ and $x^{adv}_0$ are still viewed as similar at $block\_14\_add$, we can assume that the change resides in this sequence of layers. We select all layers in block $15$ and visualize them, generating the image on the bottom half of Figure~\ref{fig:imagenet_mobilenet_1}. With this visualization, we can observe the effects of connected convolutional or dense layers and supporting layers, such as batch normalization or additional activation layers. 

\begin{figure}[!h]
    \centering
    \includegraphics[width=0.425\textwidth]{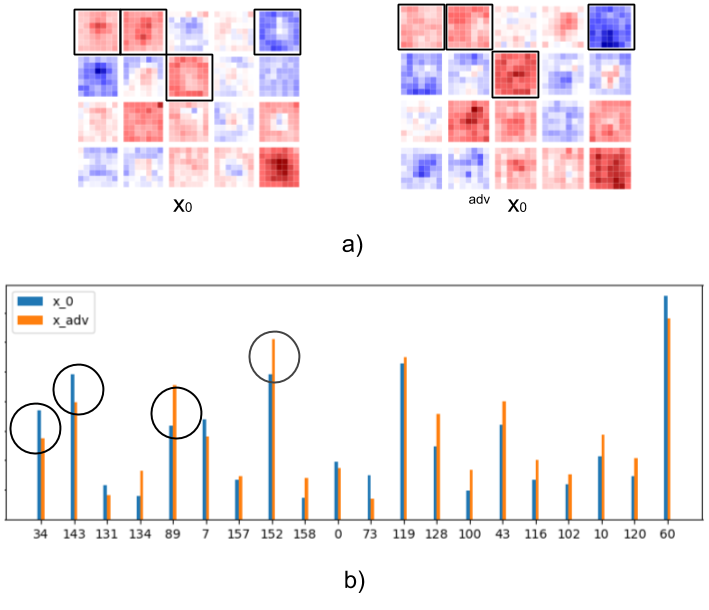}
    \caption{Analysis of outputs for $x_0$ and $x^{adv}_0$ at the layer $block\_15\_project$ of \textit{MobileNet V2}. a) shows the $20$ units with the most expressive changes in perception between the two examples (highest absolute difference). Blue and red represent positive and negative outputs, respectively. b) displays the absolute output sum for each of the $20$ units on each data example. Units $34$, $143$, $89$, and $152$ (circled in black in both images) show high overall outputs and high variations, being more likely to be relevant for classification for objects $x_0$ and $x_adv$.}
    \label{fig:imagenet_mobilenet_2}
\end{figure}

\begin{figure}[!h]
    \centering
    \includegraphics[width=0.425\textwidth]{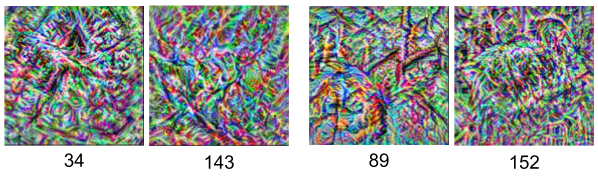}
    \caption{Image patterns detected by the four most relevant units in layer \textit{block\_15\_project}, as identified in Figure~\ref{fig:imagenet_mobilenet_2}. Correlating filters and visual features can help explain how an adversarial example is able to exploit a CNN model, especially concerning the distinction between robust and non-robust features.}
    \label{fig:imagenet_mobilenet_3}
    \vspace{-0.5cm}
\end{figure}

From Figure~\ref{fig:imagenet_mobilenet_1}, it is possible to notice a clear separation at $block\_15\_project$, so we select it for inspection using the \textit{Layer View} (Figure~\ref{fig:imagenet_mobilenet_2}). Outputs on this layer are slightly more complex to visualize due to negative values that can be carried over to the next weighted layer without passing through a rectified linear function. We build activation heatmaps in the top row by calculating the sum of all absolute output values for each unit. In these heatmaps, it is noticeable that, while changes happen between the two objects, they are in the middle of the range, while the highest outputs remain the same. We calculate the total difference between all unit outputs for $x_0$ and $x^{adv}_0$, and select the units with the $20$ most diverging values. Their activation maps are displayed in the second row, with positive outputs in blue and negative outputs in red. It is possible to notice that many of the units perceive differences between $x_0$ and $x^{adv}_0$ at the center of the picture.

We then export activation data and use a graph to correlate the output sum at each unit with units that show the most diverging outputs for the two images to find units where intense changes occur and, at the same time, produce strong outputs since large output values are most likely to be used for classification. Four of them were highlighted due to visually distinct behavior. They are circled in black, and activation maps for these units were also highlighted in black in the second upper row. Two of them ($34$, $143$) show stronger outputs for $x_0$, while the other two ($89$, $152$) show stronger outputs for $x_{adv}$. 

Using filter visualization, these units are then visualized in input space, as can be seen in the last row of Figure~\ref{fig:imagenet_mobilenet_2}. These are the visual features captured by the most divergent filters in this layer, which can then be correlated to patterns observed in the images themselves. An analysis can be performed as to their relevance in visually identifying the original images to determine the presence of robust vs. non-robust features. Model improvement regarding robust features can be achieved by performing further training using images that clearly show these patterns, while non-robust features may reveal insights on dataset bias and overfitting. 

\section{Discussion and Limitations}

Our approach's main limitation is that the effectiveness of the \textit{Flow View} representations depend on how representative the background data is. We conducted experiments with randomly sampled data from smaller training datasets such as MNIST or CIFAR10~\cite{krizhevsky2009cifar10} and empirically observed that sample sizes as small as $1,000$ instances generated \textit{Activation Profiles} similar to the complete dataset. However, for large datasets such as the \textit{ImageNet}, it is not viable to draw random samples from the whole data distribution, as discussed in Section~\ref{sec:mnet_imagenet_1}. Therefore, it is crucial to draw background data from samples that are somewhat representative of the training dataset's subdomain, whose features are being observed.

Another limitation in the \textit{Activation Profiles} is that the vertical axis carries no inherent meaning in our visual representations. Visual analysis is based on distances between points and how observed data relate to background groups. However, the fact that a specific group is displayed above or below another is simply a byproduct of the distance minimization process. It is possible to add information to this axis by controlling alignment and initialization, but it is out of this paper's scope.

Finally, our analysis over which units are exploited by an adversarial example can display locations where the greatest differences are detected, but it does not guarantee that these units are responsible for classification. It is known that the presence of high activation values does not necessarily mean that a given unit has a strong influence in model output for that object~\cite{liu2018analyzing}. However, observing why specific units display different behavior towards the adversarial example despite not being relevant to that classification might provide insights into this analysis. Combining our framework with other approaches, such as AEVis~\cite{liu2018analyzing}, could provide informative results.
\section{Conclusion}

This paper presented a novel visual analytics framework capable of summarizing model perception of input data and highlighting differences between adversarial and control examples. Users can acquire insight over areas of interest that can be observed in detail by examining hidden layer outputs and components of its inner structure, such as layer weights or convolutional filters. We show its application in two study cases, a small benchmark data/model combination and a large dataset/deep network model scenario more akin to real-world applications.

Our framework comprehends a novel visualization approach (the \textit{Activation Profiles}), as well as building upon state-of-the-art techniques to provide detailed visual information (\textit{Architecture View} and \textit{Layer View}). This framework is intended as a guide to provide insight into a model's behavior, as opposed to direct suggestions for model improvement/debugging. Its utilization can be combined with other tools for a detailed mental picture of how representations are formed within hidden layers and how adversarial examples exploit these structures to provoke unexpected outputs. 

Future work may follow different paths. The framework can be extended to provide full support for non-sequential network architectures (such as RNNs) and comparison between different models. The \textit{background flow} and \textit{class fingerprint} metaphors offer insight in understanding group data behavior, and their utility can be investigated in different contexts, such as supporting training and model topology design.

\bibliographystyle{eg-alpha} 

\bibliography{ref}
\end{document}